\documentclass{article} % For LaTeX2e
\usepackage[preprint]{colm2025_conference}

\usepackage{microtype}
\usepackage{hyperref}
\usepackage{url}
\usepackage{booktabs}

\usepackage{lineno}

\definecolor{darkblue}{rgb}{0, 0, 0.5}
\hypersetup{colorlinks=true, citecolor=darkblue, linkcolor=darkblue, urlcolor=darkblue}

\usepackage{graphicx}
\usepackage{adjustbox}
\usepackage{booktabs}
\usepackage{multirow}
\usepackage{xcolor}
\usepackage{arydshln}
\usepackage{amssymb} % For arrows
\usepackage{nicematrix}
\usepackage{graphicx}

\newcommand{\ctslogo}{\raisebox{3.4pt}{\includegraphics[scale=0.01]{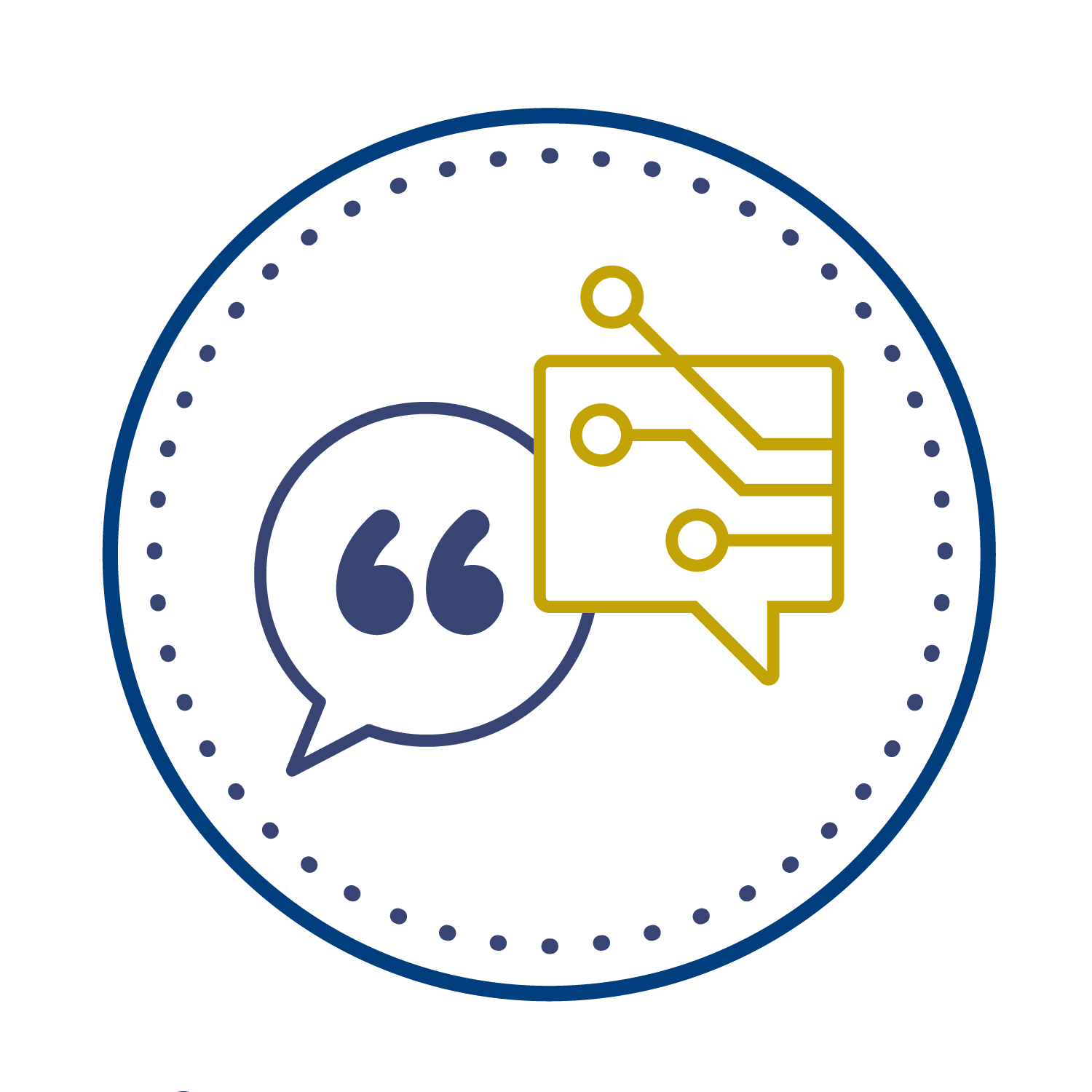}}}
\newcommand{\PAIlogo}{\raisebox{3.4pt}{\includegraphics[scale=0.08]{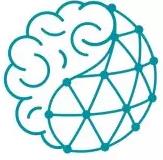}}}
\newcommand{\uoslogo}{\raisebox{3.2pt}{\includegraphics[scale=0.065]{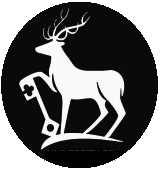}}}

\title{ALOPE: Adaptive Layer Optimization for Translation Quality Estimation using Large Language Models}

\author{Archchana Sindhujan\PAIlogo , Shenbin Qian\ctslogo, Chan Chi Chun Matthew\uoslogo , \\
\bf{Constantin Or\u{a}san\ctslogo \& Diptesh Kanojia\PAIlogo}\\
\\
\PAIlogo Institute for People-Centred AI and \ctslogo Centre for Translation Studies,\\ 
\uoslogo School of Computer Science and Electronic Engineering,\\
\uoslogo University of Surrey, United Kingdom\\
\\
\texttt{\{a.sindhujan, s.qian, c.orasan, d.kanojia\}@surrey.ac.uk},\\
\texttt{matthewchancc@gmail.com}
}

\begin{document}

\ifcolmsubmission
\linenumbers
\fi

\maketitle

\begin{abstract}
Large Language Models (LLMs) have shown remarkable performance across a wide range of natural language processing tasks. Quality Estimation (QE) for Machine Translation (MT), which assesses the quality of a source-target pair without relying on reference translations, remains a challenging cross-lingual task for LLMs. The challenges stem from the inherent limitations of existing LLM-based QE systems, which are pre-trained for causal language modelling rather than regression-specific tasks, further elevated by the presence of low-resource languages given pre-training data distribution. This paper introduces ALOPE, an adaptive layer-optimization framework designed to enhance LLM-based QE by restructuring Transformer representations through layer-wise adaptation for improved regression-based prediction. Our framework integrates low-rank adapters (LoRA) with \textit{regression task heads}, leveraging selected pre-trained Transformer layers for improved cross-lingual alignment. In addition to the layer-specific adaptation, ALOPE introduces two strategies—\textit{dynamic weighting}, which adaptively combines representations from multiple layers, and \textit{multi-head regression}, which aggregates regression losses from multiple heads for QE. Our framework shows improvements over various existing LLM-based QE approaches. Empirical evidence suggests that intermediate Transformer layers in LLMs provide contextual representations that are more aligned with the cross-lingual nature of the QE task. We make resultant models and framework code publicly available\footnote{\href{https://github.com/surrey-nlp/ALOPE}{https://github.com/surrey-nlp/ALOPE}} for further research, also allowing existing LLM-based MT frameworks to be scaled with QE capabilities.

\end{abstract}

\section{Introduction}

Quality Estimation (QE) allows the output of Machine Translation (MT) to be evaluated at scale without the need for reference translations~\citep {zerva-etal-2022-findings}. MT QE can be performed at different levels, but this paper focuses on segment-level evaluation for low-resource languages, predicting the Direct Assessment (DA) score ($0 \leq x \leq 100$) for each segment~\citep{graham-etal-2013-continuous}. The ground truth score is obtained by averaging scores assigned by multiple human annotators ($>=3$) following annotation guidelines that emphasize adequacy, fluency, and \textit{cross-lingual meaning transfer} within source and machine-translated output. 

Although QE annotation frameworks such as Multi-dimensional Quality Metrics (MQM)~\citep{burchardt-2013-multidimensional} provide fine-grained insights into translation errors, they impose a considerable cognitive burden on annotators, often making them more labour-intensive. In contrast, DA scores offer a more efficient alternative for complementing such frameworks in the overall evaluation of segment-level MT quality. Furthermore, DA scores can be leveraged to guide LLMs in reasoning over translation errors and in identifying translations that require human inspection or post-editing.

While Large Language Models (LLMs) exhibit strong performance in translation evaluation when reference translations are available~\citep{kocmi-federmann-2023-large,qian-etal-2024-large}, assessing cross-lingual meaning transfer in a reference-less setting remains a significant challenge~\citep{vandan-etal-2023-towards,sindhujan-etal-2025-llms} for LLMs. Due to the limited representation of languages in pre-training data~\citep{zhu2024multilinguallargelanguagemodels}, complexity of translation errors~\citep{sindhujan2024optimizing}, and cross-lingual alignment within LLMs~\citep{nguyen-etal-2024-democratizing}, LLM-based QE does not perform as well as state-of-the-art (SoTA) encoder-based models like CometKiwi~\citep{rei-etal-2023-scaling}.

The prediction of DA scores for QE constitutes a regression task, requiring fine-grained numerical predictions. However, instructing LLMs to predict a DA score through prompt engineering or instruction tuning for the generation task lacks regression-specific training objectives. This limitation often results in outputs with reduced granularity, thereby constraining their effectiveness in tasks like QE that demand precise scalar predictions~\citep{zerva-etal-2024-findings}. Furthermore, current LLM-based methods for cross-lingual tasks such as QE predominantly rely on representations from only the final Transformer layer, overlooking potentially valuable intermediate layer embeddings. Recent findings by \citet{kargaran-etal-2025-mexa} indicate that optimal cross-lingual alignment is not necessarily achieved at the final Transformer layer for low-resource languages.

To address these challenges, we introduce the Adaptive Layer Optimization for Translation Quality Estimation (ALOPE) framework, a novel approach that integrates dedicated regression heads and low-rank adapters (LoRA) within Transformer layers, facilitating efficient instruction-based fine-tuning with LLMs. Our framework systematically investigates Transformer layers with regression heads, enabling the identification of optimal configurations that maximize cross-lingual transfer learning for segment-level translation quality estimation. Beyond investigating layer-wise embeddings, ALOPE introduces two additional strategies—dynamic weighting and multi-head regression, which leverage multiple Transformer layers for QE. Moreover, the framework is developed to be model-agnostic, enabling seamless integration of any pre-trained LLMs with minimal configuration, thus promoting ease of deployment and scalability.

We evaluate ALOPE by comparing its performance against zero-shot inference and standard instruction fine-tuned (SIFT) LLMs across four open-source models ($\leq$8B parameters) on eight low-resource language pairs. The results highlight that ALOPE consistently outperforms SIFT, demonstrating its effectiveness by identifying optimal Transformer layers with better cross-lingual transfer learning, which benefits the quality estimation. The main contributions of this paper are,

\begin{itemize}

\item We propose ALOPE, an adaptable framework that integrates regression heads with LoRA, enabling efficient QE with any LLMs by facilitating optimal cross-lingual representations.

\item ALOPE consistently outperforms the best results achieved by standard instruction fine-tuned LLMs across all eight low-resource language pairs evaluated.

\item ALOPE introduces two additional approaches, dynamic-weighting and multi-layer regression, which utilize combined Transformer layers strategies and outperform the LLMs baseline under most conditions evaluated. 

\end{itemize}

% \vspace{-10pt}

\section{Background } \label{sec:background}

In recent years, significant advancements have been made in the field of quality estimation. The development of QE models has evolved from early feature engineering-based approaches~\citep{specia2013quest,Specia2015} to more sophisticated methodologies, progressing through basic neural networks~\citep{kim2017predictor} and deep neural network architectures~\citep{ive2018deepquest,kepler2019openkiwi}. Currently, Transformer-based architectures represent the state-of-the-art in QE~\citep{blain-etal-2023-findings,rei-etal-2022-cometkiwi,ranasinghe-etal-2020-transquest,perrella2022matese,moura2020unbabel,baek2020patquest}.

Recent advancements have positioned LLMs as a powerful tool across various NLP tasks~\citep{zhao2023survey}. In the context of translation evaluation,~\citet{kocmi-federmann-2023-large} introduced the GEMBA metric, using zero-shot prompting across nine GPT variants and four prompt types for three language pairs. While GEMBA achieved best results with reference translations, its performance in referenceless QE remained below SOTA, highlighting the continued challenges LLMs face in reference-free evaluation settings. ~\citet{vandan-etal-2023-towards} investigated QE for low-resource languages by pre-tuning adapters on a large English-Indic parallel corpus for machine translation and subsequently fine-tuning the model with supervised QE data. Their findings suggest that pre-tuning with machine translation data does not enhance QE performance for low-resource languages and that LLMs struggle to transfer MT knowledge effectively for QE tasks in these settings.  

It has been noted that LLM-based approaches continue to underperform compared to traditional predictor-estimator architectures in QE, particularly for sentence-level scoring~\citep{zerva-etal-2024-findings}. Predictor-estimator models, which explicitly treat QE as a regression task, provide finer-grained quality assessments by effectively distinguishing between varying translation qualities~\citep{kim2017predictor}. In contrast, LLMs, which rely on prompt engineering or instruction tuning, tend to produce outputs with limited granularity, often defaulting to narrower scoring ranges. This lack of differentiation restricts their ability to model nuanced quality variations across translations, contributing to their comparatively weaker performance in QE~\citep{kocmi-federmann-2023-large, vu-etal-2024-foundational}.

Cross-lingual transfer learning is intended to improve low-resource language performance by transferring knowledge from high-resource languages through shared linguistic features~\citep{choenni-etal-2023-cross, pham-etal-2024-unibridge}. However, despite its promise, existing studies show that the approach has so far been more effective in high-resource settings~\citep{zhu2024multilinguallargelanguagemodels}, while its benefits for low-resource languages remain limited due to data scarcity and linguistic diversity~\citep{zhu2024multilinguallargelanguagemodels}. The work of~\citet{sindhujan-etal-2025-llms} reinforces the cross-lingual challenge in QE with LLMs, highlighting the poor tokenization issue for low-resource languages. Nonetheless, optimizing cross-lingual transfer remains essential for improving LLM-based QE.

% \vspace{-5pt}
\section{Methodology}\label{sec:methodolgy}

\subsection{Dataset}\label{subsec:data}

Our study mainly focuses on the QE dataset, which is derived from the WMT QE shared task~\citep{zerva-etal-2022-findings,blain-etal-2023-findings}. It contains source and translation pairs with human-annotated DA scores\footnote{Mean DA score annotations for Indic-language pair test sets can be shared upon request.}, which measure the quality of the machine translations. The QE dataset spans eight low-resource language pairs: English to \{Gujarati, Hindi,  Marathi, Tamil, Telugu\}, and \{Nepali, Estonian, Sinhala\} to English. Hindi and Estonian, while considered mid-resource languages for machine translation~\citep{nguyen-etal-2024-democratizing}, have limited resources available for quality estimation. We utilize the dataset as train and test splits, as detailed in Appendix~\ref{app:dataset}.

 \begin{figure*}[t]
 \centering
 \includegraphics[width=0.9\textwidth, keepaspectratio]{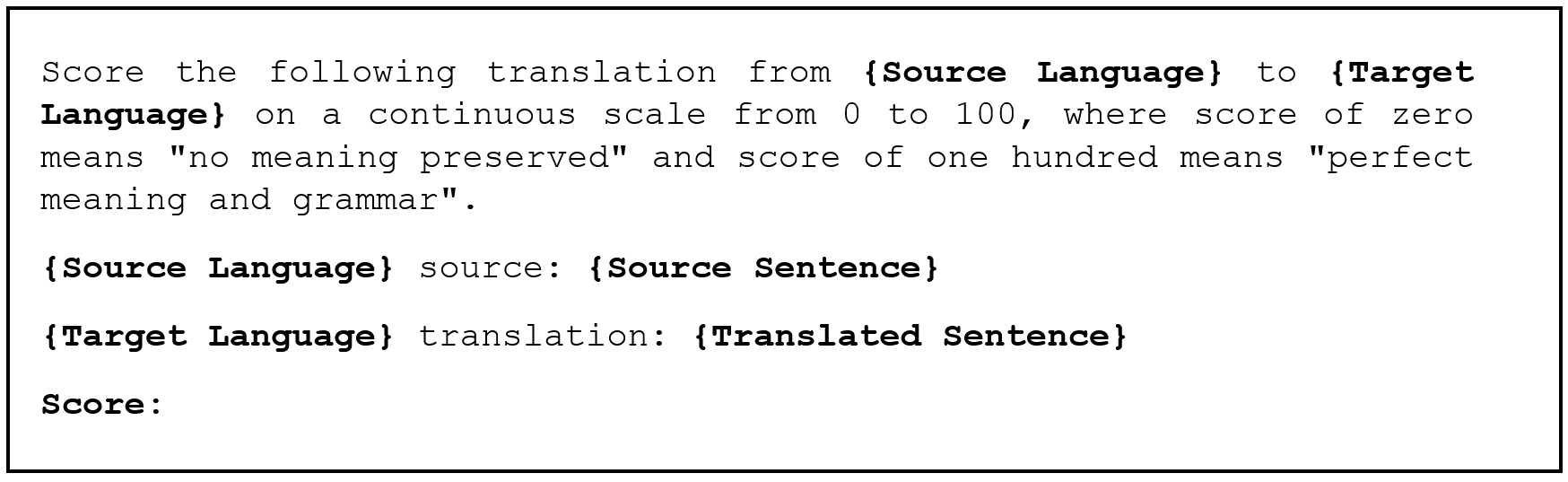}
\caption{ GEMBA prompt for quality estimation~\citep{kocmi-federmann-2023-large}}
\label{fig:gemba} % Give a unique label
\end{figure*}

\subsection{Experimental settings}\label{sec:experimental_setting}
% \vspace{-5pt}
We utilized the GEMBA prompt, which is a simple and straightforward prompt proposed by~\citet{kocmi-federmann-2023-large} for estimating the quality of the machine-translated text as shown in Figure ~\ref{fig:gemba}.
%LLMs
All our experiments were conducted with open-source generative LLMs which have 8B parameters or less: LLaMA2-7B, LLaMA3.1-8B, LLaMA3.2-3B and Aya-expanse-8B \footnote{
\href{https://huggingface.co/meta-llama/LLaMA-2-7b-chat-hf}{LLaMa-2-7B,}

\href{https://huggingface.co/meta-llama/Llama-3.1-8B-Instruct}{LLAMA3.1-8B,}
\href{https://huggingface.co/meta-llama/Llama-3.2-3B-Instruct}{LLAMA3.2-3B,}
\href{https://huggingface.co/CohereForAI/aya-expanse-8b}{Aya-expanse-8B}
}. 
Our experiments were conducted under three settings: zero-shot, standard instruction fine-tuning (SIFT) and ALOPE framework. In zero-shot experiments, we obtain the results directly from the pre-trained LLMs using vLLM framework~\citep{kwon2023efficient} and we have utilized the LLaMA-Factory framework~\citep{zheng2024llamafactory} to perform SIFT. 

All instruction-fine-tuning experiments (SIFT and ALOPE) with QE data were conducted using integrated multilingual language-pair training, which involved combining training data from the eight low-resource language pairs mentioned in section~\ref{subsec:data}. Inference was performed on language-specific test sets to evaluate model performance.

\subsection{ALOPE framework} \label{regression-head}  

LLMs are optimized for next-token prediction, excelling at generative tasks but struggling with regression-based goals like quality estimation~\citep{sindhujan-etal-2025-llms}. This limits their ability to capture fine-grained relationships between language features and numerical scores. Although LLMs refine context through multiple Transformer layers, standard practice usually predicts with the final layer to obtain the outputs.

Addressing these challenges, ALOPE enables LLMs to perform regression by integrating a regression head within the low-rank adapted Transformer architecture. We analyze the impact of adaptive regression heads to determine which Transformer layers contribute most to enhancing the performance of LLMs for cross-lingual QE tasks with low-resource language pairs.

\begin{figure*}[t]
\centering
\includegraphics[width=0.99\textwidth, keepaspectratio]{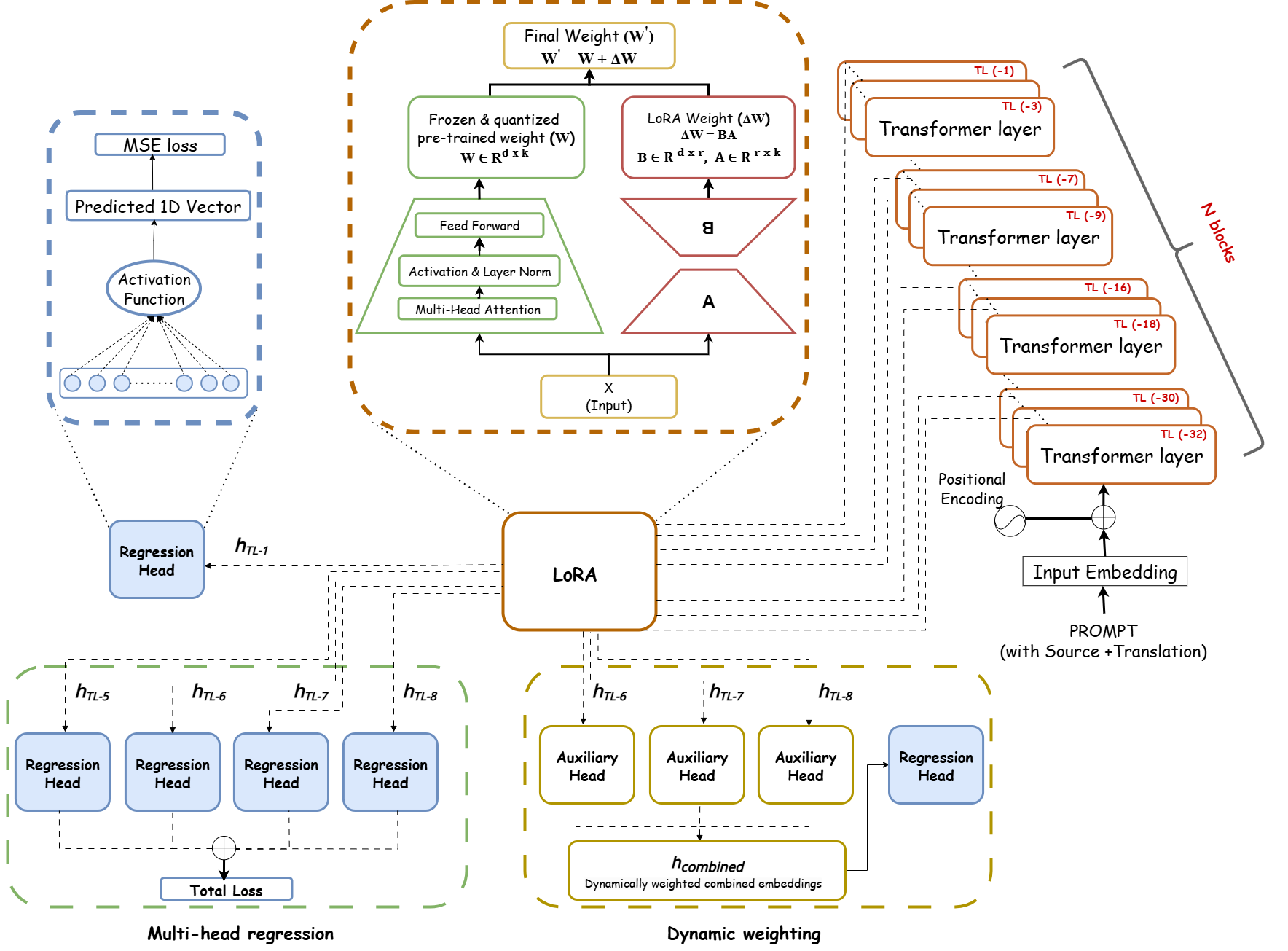}
\caption{\small{Presents the ALOPE framework. Regression heads can obtain the embeddings from any LoRA-adapted Transformer layers, enabling flexible adaptation to perform QE.} }
\label{fig:RH} % Give a unique label
\end{figure*}

We employed LoRA (Low-Rank Adaptation)~\citep{hu2022lora} for efficient fine-tuning, targeting the projection layers of the Transformer architecture. Rather than updating the full parameter set during fine-tuning, we introduce trainable low-rank matrices into the attention mechanism, allowing the model to instruction fine-tune for the QE task while keeping the original weights frozen as shown in   Figure~\ref{fig:RH}.  The adaptation to a weight matrix \( W \) is represented as \( W' = W + BA \), where \( A \in \mathbb{R}^{r \times d} \) and \( B \in \mathbb{R}^{d \times r} \) are the trainable low-rank matrices and \( r \ll d \),  where \( r \) (the rank) is much smaller than the dimension \( d \) which ensures a significant reduction in the number of tunable parameters. In our experiments, the rank \( r \) is set to 32. To further optimize memory usage and computational efficiency, we combine LoRA with 4-bit quantization~\citep{dettmers2023qlora}.
% \vspace{-4pt}

To enable regression-based prediction with LLMs, the ALOPE framework assigns a dedicated regression head integrated into the Transformer architecture. This component is designed to extract hidden representations from any selected LoRA-adapted Transformer layer, thereby facilitating quality estimation using layer-specific contextual embeddings. The hidden state extracted from the final token of the sequence from the chosen LoRA-adapted Transformer layer is passed through a regression head as shown in Figure \ref{fig:RH}. 
Let \( h_k \) be the hidden state of the final token:  
\vspace{-6pt}
\[
\mathbf{h}_k = \mathbf{H}_k[-1] \in \mathbb{R}^d
\]
where \( k \) is the chosen Transformer layer, \( d \) is the dimensionality of the hidden state,  \( \mathbf{H}_k \) represents the sequence of hidden states at layer \( k \) and \([-1]\) represents the index of the final token in the sequence at Transformer layer \( k \). The regression head consists of a single linear layer that maps the high-dimensional hidden state \( h_k \) to the final scalar output. The linear activation function is applied implicitly, as the linear layer performs an affine transformation:
\[
\hat{y} = h_k W_n, \quad \hat{y} \in \mathbb{R}
\]
where \( W_n \in \mathbb{R}^{d \times 1} \) is the weight matrix of the linear layer. The output \( \hat{y} \) is a single scalar value per sequence, representing the predicted quality score. To evaluate the model, the mean squared error (MSE) loss is computed by measuring the average squared difference between the predicted values in the 1D vector and the actual target values.

The regression-specific adaptation with regression heads and LoRA-based fine-tuning can be considered independently. However, the focus of our proposed method is efficiency in computationally resource-constrained scenarios. We specifically chose regression heads incorporated with LoRA for its parameter efficiency, reducing the computational overhead compared to full fine-tuning, and allowing modularity given a pre-hosted LLM. ALOPE is able to perform QE with LoRA-less regression heads too. A publicly available library was adopted for our implementation \footnote{ \href{https://github.com/center-for-humans-and-machines/Transformer-heads/tree/main}{https://github.com/center-for-humans-and-machines}}. 

In the ALOPE framework, the regression heads are assigned numerical indices, where negative values indicate Transformer layers indexed in reverse from the final layer. For example, as shown in the Figure~\ref{fig:RH}, placing a regression head at layer -1 (TL-1) corresponds to extracting the hidden state from the final Transformer layer, while layer-4 (TL-4) refers to the layer located four positions before the last. This indexing approach provides a consistent reference for selecting layers across different models with different numbers of Transformer layers. To identify the optimal Transformer layer ALOPE systematically explores six candidate Transformer layers across all evaluated LLMs: the final layer (-1), intermediate layers (-7 and -11), a mid-range layer (-16), and lower-level layers (-20 and -24).

In addition to the layer-specific regression head approach (vanilla ALOPE), we extend the ALOPE framework with two additional strategies—dynamic weighting and multi-head regression, both of which leverage multiple Transformer layers to enhance translation quality estimation through multiple layer-level representation modelling.

\subsubsection{Dynamic weighting}
In the dynamic weighting approach, we extract sequence-level contextual representations from selected Transformer layers using auxiliary heads. Auxiliary heads are solely utilized to extract embeddings from specific Transformer layers and are not subjected to direct supervised training. To obtain a sequence-level embedding from the Transformer layer \( k \), we select the final token's hidden state \(h_k \). Instead of treating these selected layers equally, we assign a trainable scalar weight \( w_k \) to each layer \( k \), and normalize these weights using the softmax function to identify each layer’s contribution:

\[
\alpha_k = \frac{\exp(w_k)}{\sum_{j=1}^{L} \exp(w_j)} \quad \text{for } k = 1, \dots, L
\]
These weights are initialized uniformly and subsequently optimized during training via backpropagation. The final embedding is a weighted sum of the individual layer embeddings:
% \vspace{-0.9ex} % Try reducing gap right after the table
% \noindent
% \vspace{-10pt}
\[
\mathbf{h}_{\text{combined}} = \sum_{k=1}^{L} \alpha_k h_k
\]
% \vspace{-1.5pt}
This representation is passed through a regression head to produce the predicted translation quality score. The model is optimized by minimizing the mean squared error between the predicted score and the ground truth. We refer to this method as \textit{dynamic weighting} because it allows the model to dynamically learn to prioritize information from the most informative layers for the task of translation quality estimation.
% \vspace{-5pt}

\subsubsection{Multi-head regression}
In this approach, multiple regression heads are integrated into different Transformer layers, each tasked with independently approximating translation quality. To ensure a balanced optimization, a weighted aggregation of individual regression head losses is employed before back-propagation. The computed final loss is back-propagated through the model and updates the selected heads. This approach enables simultaneous learning across multiple layers. By optimizing a combined loss across selected layers, the framework effectively integrates multi-layered contextual information. During inference, the final translation quality prediction is derived by averaging the aggregated outputs from these regression heads.

% \subsubsection{Combined loss}

\begin{table*}[t]
\centering
\small
\resizebox{0.98\textwidth}{!}{
\begin{NiceTabular}{c|cccccc:ccc|c}
\midrule
  & \textbf{Model} & \textbf{En-Gu} & \textbf{En-Hi} & \textbf{En-Mr} & \textbf{En-Ta} & \textbf{En-Te} & \textbf{Et-En} & \textbf{Ne-En} & \textbf{Si-En} & \textbf{\textit{Avg}} \\
\midrule

\parbox[t]{2mm} {\multirow{5}{*}{ \centering \rotatebox[origin=c]{90}{ \textbf{\small {Zero-shot}}}}}
& LLaMA2-7B & 0.006 & -0.002 & 0.053 & 0.067 & -0.016 & 0.168 & 0.153 & 0.144 & 0.072 \\
 & LLaMA3.1-8B & 0.164 & 0.194 & 0.245 & 0.220 & 0.132 & 0.503 & 0.262 & 0.267 & \textit{0.248} \\
 & LLaMA3.2-3B & 0.095 & 0.095 & 0.116 & 0.128 & 0.098 & 0.359 & 0.216 & 0.120 & \textit{0.153} \\
 & Aya-expanse-8B & 0.123 & 0.135 & 0.089 & 0.117 & 0.049 & 0.315 & 0.352 & 0.330 & \textit{0.189} \\
 \\
 & {\textbf{\textit{Avg}}} & \textit{0.097} & \textit{0.106} & \textit{0.126} & \textit{0.133} & \textit{0.066} & \textit{0.336} & \textit{0.246} & \textit{0.215} &  \\

\midrule
% \multirow{7}{*}{\centering \rotatebox[origin=c]{90}{ \textbf{\scriptsize {TL (-1)}}}}
\parbox[t]{2mm} {\multirow{5}{*}{\centering \rotatebox[origin=c]{90}{ \textbf{\small {TL (-1)}}}}}

& LLaMA2-7B & 0.563~$\uparrow$ & 0.414~$\uparrow$ &  \textsuperscript{*}0.609~$\uparrow$ & 0.525~$\uparrow$ & 0.356~$\uparrow$ &  \textsuperscript{*}0.742~$\uparrow$ & 0.596~$\uparrow$ & \textsuperscript{*}0.565~$\uparrow$ & 0.546\\
& LLaMA3.1-8B & \textsuperscript{*}0.594~$\uparrow$ &  \textsuperscript{*}0.469~$\uparrow$ &  \textsuperscript{*}0.620~$\uparrow$ & 0.567 & 0.363~$\uparrow$ & 0.734~$\uparrow$ & 0.647~$\uparrow$ & 0.547 & \textit{0.567} \\
& LLaMA3.2-3B & \textsuperscript{*}0.604~$\uparrow$ &  \textsuperscript{*}0.477~$\uparrow$ & \textbf{0.636}~$\uparrow$ & 0.580~$\uparrow$ & 0.348~$\uparrow$ & 0.735~$\uparrow$ & \textsuperscript{*}0.674~$\uparrow$ & 0.543~$\uparrow$  & \textit{0.575}\\
& Aya-expanse-8B & 0.068 & 0.178 & 0.219 & -0.006 & 0.275 & 0.115 & 0.012 & 0.077 & \textit{0.117}\\
\\
& {\textbf{\textit{Avg}}} & \textit{0.457} & \textit{0.385} & \textit{0.521} & \textit{0.416} & \textit{0.335} & \textit{0.581} & \textit{0.482} & \textit{0.433} \\

\midrule

\parbox[t]{2mm} {\multirow{5}{*}{\centering \rotatebox[origin=c]{90}{ \textbf{\small {TL (-7)}}}}}

& LLaMA2-7B & 0.567~$\uparrow$ & 0.336~$\uparrow$ & 0.542~$\uparrow$ & 0.484~$\uparrow$ & 0.317~$\uparrow$ &  \textsuperscript{*}0.739~$\uparrow$ & 0.606~$\uparrow$ & \textbf{0.573}$\uparrow$ & 0.520\\
& LLaMA3.1-8B &  \textsuperscript{*}0.590~$\uparrow$ &  \textsuperscript{*}0.477~$\uparrow$ &  \textsuperscript{*}0.625~$\uparrow$ & 0.528 & \textbf{0.388}~$\uparrow$ &  \textsuperscript{*}0.744~$\uparrow$ & 0.638~$\uparrow$ & 0.544 & \textit{0.567} \\
& LLaMA3.2-3B & \textbf{0.606}~$\uparrow$ & \textbf{0.479}$\uparrow$ &  \textsuperscript{*}0.617$\uparrow$ & 0.585$\uparrow$ &  \textsuperscript{*}0.369$\uparrow$ &  \textbf{0.751}$\uparrow$ & 0.664$\uparrow$ & \textsuperscript{*}0.553$\uparrow$ & \textit{0.578}\\
& Aya-expanse-8B & 0.538~$\uparrow$ & 0.447~$\uparrow$ & 0.597~$\uparrow$ & 0.528~$\uparrow$ & 0.347~$\uparrow$ &  \textsuperscript{*}0.741~$\uparrow$ & 0.646~$\uparrow$ & 0.544~$\uparrow$ & \textit{0.549}\\
% \hline 
\\

& {\textbf{\textit{Avg}}} & \textit{0.575}\textsuperscript{$\dagger$} & \textit{0.435}\textsuperscript{$\dagger$} & \textit{0.595}\textsuperscript{$\dagger$} & \textit{0.531}\textsuperscript{$\dagger$} & \textit{0.355}\textsuperscript{$\dagger$} & \textit{0.744}\textsuperscript{$\dagger$} & \textit{0.639}\textsuperscript{$\dagger$} & \textit{0.554}\textsuperscript{$\dagger$} \\

\midrule
% \multirow{3}{*}{\textbf{Layer (-11)}} 
% \multirow{3}{*}{\rotatebox[origin=c]{90}{ \textbf{\scriptsize {TL (-11)}}}}
\parbox[t]{2mm} {\multirow{5}{*}{\centering \rotatebox[origin=c]{90}{ \textbf{\small {TL (-11)}}}}}

& LLaMA2-7B & 0.360 & 0.301 & 0.361 & 0.254 & 0.293$\uparrow$ & 0.405 & 0.164 & 0.049 & 0.273 \\
& LLaMA3.1-8B & 0.514 & 0.412$\uparrow$ &  \textsuperscript{*}0.609$\uparrow$ & 0.438 & 0.304$\uparrow$ & 0.148 & 0.554 & 0.493 & \textit{0.434}\\
& LLaMA3.2-3B & \textsuperscript{*}0.594$\uparrow$ &  \textsuperscript{*}0.476$\uparrow$ & 0.605$\uparrow$ & \textbf{0.610}$\uparrow$ &  \textsuperscript{*}0.373$\uparrow$ &  \textsuperscript{*}0.748$\uparrow$ & \textsuperscript{*}0.678$\uparrow$ & \textsuperscript{*}0.560$\uparrow$ & \textit{0.581}\\

& Aya-expanse-8B & 0.490 & 0.411$\uparrow$ & 0.572$\uparrow$ & 0.445 & 0.336$\uparrow$ & 0.569 & 0.453 & 0.439$\uparrow$ & \textit{0.464}\\
\\
& {\textbf{\textit{Avg}}} & \textit{0.489} & \textit{0.400} & \textit{0.537} & \textit{0.437} & \textit{0.327} & \textit{0.467} & \textit{0.462} & \textit{0.385} \\

\midrule

\parbox[t]{2mm} {\multirow{5}{*}{\centering \rotatebox[origin=c]{90}{ \textbf{\small {TL (-16)}}}}}

& LLaMA2-7B & 0.540~$\uparrow$ & 0.381~$\uparrow$ & 0.585~$\uparrow$ & 0.482~$\uparrow$ & 0.308~$\uparrow$ & \textsuperscript{*}0.751~$\uparrow$ & 0.580~$\uparrow$ & \textsuperscript{*}0.569~$\uparrow$ & 0.524 \\
& LLaMA3.1-8B & 0.558~$\uparrow$ & 0.453~$\uparrow$ & 0.602~$\uparrow$ & 0.523 & 0.350~$\uparrow$ & \textsuperscript{*}0.737~$\uparrow$ & 0.652~$\uparrow$ & 0.513 & \textit{0.548}\\
& LLaMA3.2-3B & 0.557~$\uparrow$ &  \textsuperscript{*}0.459~$\uparrow$ & 0.597~$\uparrow$ & 0.547 & 0.338~$\uparrow$ & \textsuperscript{*}0.745~$\uparrow$ & \textbf{0.682}$\uparrow$ & \textsuperscript{*}0.567~$\uparrow$ & \textit{0.561}\\
& Aya-expanse-8B & 0.467 & 0.390~$\uparrow$ & 0.557~$\uparrow$ & 0.481 & 0.314~$\uparrow$ & 0.727~$\uparrow$ & 0.576~$\uparrow$ & 0.540~$\uparrow$ & \textit{0.506} \\

% \hline 
\\
& {\textbf{\textit{Avg}}} & \textit{0.530} & \textit{0.421} & \textit{0.585} & \textit{0.508} & \textit{0.327} & \textit{0.740} & \textit{0.622} & \textit{0.547} \\

\midrule

\parbox[t]{2mm} {\multirow{5}{*}{\centering \rotatebox[origin=c]{90}{ \textbf{\small {TL (-20)}}}}}

% \multirow{3}{*}{\rotatebox[origin=c]{90}{ \textbf{\scriptsize {TL (-20)}}}}
& LLaMA2-7B & 0.470$\uparrow$ & 0.405$\uparrow$ & 0.544$\uparrow$ & 0.460$\uparrow$ & 0.338$\uparrow$ & 0.684$\uparrow$ & 0.508$\uparrow$ & 0.534$\uparrow$ & 0.493 \\

& LLaMA3.1-8B & 0.484 & 0.394$\uparrow$ & 0.553$\uparrow$ & 0.321 & 0.172 & 0.649 & 0.524 & 0.494 & \textit{0.449}\\
& LLaMA3.2-3B & 0.430 & 0.408$\uparrow$ & 0.579$\uparrow$ & 0.303 & 0.286$\uparrow$ & 0.601 & 0.488 & 0.464$\uparrow$ & \textit{0.445}\\
& Aya-expanse-8B & 0.437 & 0.300 & 0.488 & 0.263 & 0.287 & 0.483 & 0.438 & 0.395$\uparrow$ & \textit{0.386}\\
\\

& {\textbf{\textit{Avg}}} & \textit{0.455} & \textit{0.377} & \textit{0.541} & \textit{0.337} & \textit{0.271} & \textit{0.604} & \textit{0.490} & \textit{0.472} \\

\midrule

% \multirow{3}{*}{\rotatebox[origin=c]{90}{ \textbf{\scriptsize {TL (-24)}}}}
\parbox[t]{2mm} {\multirow{5}{*}{\centering \rotatebox[origin=c]{90}{ \textbf{\small {TL (-24)}}}}}
& LLaMA2-7B & 0.500$\uparrow$ & 0.421$\uparrow$ & 0.538$\uparrow$ & 0.379 & 0.239 & 0.630$\uparrow$ & 0.507$\uparrow$ & 0.472$\uparrow$ & 0.461 \\

& LLaMA3.1-8B & 0.421 & 0.378$\uparrow$ & 0.552$\uparrow$ & 0.330 & 0.290$\uparrow$ & 0.515 & 0.530 & 0.464 & \textit{0.435}\\
& LLaMA3.2-3B & 0.443 & 0.376 & 0.507 & 0.367 & 0.299$\uparrow$ & 0.559 & 0.528 & 0.487$\uparrow$ & \textit{0.446}\\
& Aya-expanse-8B & 0.375 & 0.319 & 0.440 & 0.337 & 0.220 & 0.393 & 0.407 & 0.345 & \textit{0.354}\\
\\
& {\textbf{\textit{Avg}}} & \textit{0.435} & \textit{0.373} & \textit{0.509} & \textit{0.353} & \textit{0.262} & \textit{0.524} & \textit{0.493} & \textit{0.442} \\

\bottomrule
\end{NiceTabular}
}
\caption{\small{Spearman correlation scores for zero-shot inference and ALOPE experiments across different Transformer layers (TL). 
TL(-1) refers to the final Transformer layer, TL(-7), TL(-11) are intermediate layers, TL(-16) is mid-layer, and TL(-20), TL(-24) are lower-level layers. $\uparrow$ indicates cases where ALOPE outperforms standard instruction fine-tuning (SIFT) for a given language pair (Correlation scores for SIFT can be found in Appendix~\ref{app:finetune}). Bolded values denote the highest correlation scores per language pair; (\textsuperscript{$\dagger$}) marks the highest average across Transformer layers; (\textsuperscript{*}) indicates the statistically insignificant Spearman scores compared to the best scores. The dashed line separates language pairs where English is the target language. } }
\label{tab:best-rh-layers_zero-shot}
\end{table*}
\vspace{-5pt}

\subsection{Evaluation \& Metrics}

We use Spearman's correlation ~\citep{sedgwick2014spearman} between the DA mean (averaged human-annotated DA scores from three or more annotators) and predictions as our evaluation metric. Additionally, we performed the Williams significance test~\citep{graham-baldwin-2014-testing, williams1959regression} to assess whether the top-performing model for each language pair achieved a significantly higher Spearman correlation than other models.
% \vspace{-10pt}

\section{Results and Discussion}\label{sec:results_discussion}
% \vspace{-5pt}

Table~\ref{tab:best-rh-layers_zero-shot} reports the Spearman correlation scores obtained under zero-shot evaluation, alongside the results from the ALOPE framework with regression heads placed at various Transformer layers (See section ~\ref{regression-head} ). The table also highlights cases where ALOPE yields improvements over standard instruction fine-tuning (SIFT) results with LLMs. Notably, the performance under zero-shot settings is substantially lower across all eight low-resource language pairs when compared to both SIFT and ALOPE. When benchmarked against the best Spearman scores from SIFT (refer to Appendix~\ref{app:finetune}), ALOPE obtains the best correlation scores for all evaluated language pairs. 
% \vspace{-5pt}
\subsection{Layer-specific adaptation for QE}\label{layerwise_analysis}
% \vspace{-5pt}

% --Discuss the best scores and significance score how layers differ and what could be the finding--- 
The experimental results obtained using the ALOPE framework, presented in Table~\ref{tab:best-rh-layers_zero-shot}, reveal that the most effective performance across language pairs is not achieved when the regression head is placed at the final Transformer layer (TL-1), but rather when it is positioned at intermediate layers (TL-7, TL-11) between the final and mid-level layers. 

Among all layers evaluated, TL-7 consistently yielded the highest Spearman correlation scores for the majority of language pairs, demonstrating both top-performing outcomes and a high number of instances with statistically insignificant differences from the best scores. The layer-wise averages shown in Table~\ref{tab:best-rh-layers_zero-shot} also confirm that TL-7 consistently offers the highest average correlation across models and language pairs. These suggest that TL-7 provides robust and reliable contextual representations suitable for quality estimation of low-resource language pairs. 

Additionally, TL-11, while producing the highest score for En-Ta, exhibited statistically insignificant differences from the best-performing layer across the other seven language pairs. This further reinforces the effectiveness of intermediate layers in capturing cross-lingual alignment.  Collectively, these results indicate that optimal layer selection is a key factor in enhancing QE performance with LLMs and that intermediate Transformer layers are better suited for extracting representations that support accurate quality prediction in low-resource machine translation tasks.

At the mid-layer (TL-16), the Ne-En language pair achieved the highest correlation, while three other language pairs (En-Hi, Et-En, and Si-En) exhibited performance that was statistically indistinguishable from the best scores, as shown in Table~\ref{tab:best-rh-layers_zero-shot}. Notably, three out of these four language pairs share a common characteristic: English serves as the target language in the translation direction. This pattern suggests that cross-lingual transfer learning in Transformer-based LLMs reaches representational consistency earlier from the mid-layer onwards, when English is the target language. The earlier stabilization may be attributed to the extensive exposure of LLMs to English during pre-training, which enhances cross-lingual alignment when generating English translations~\citep{kargaran2024mexamultilingualevaluationenglishcentric}. Furthermore, as shown in Table~\ref{tab:best-rh-layers_zero-shot}, models consistently achieve higher average correlation scores when English is the target language in MT, indicating stronger cross-lingual transfer in that direction. Notably, Et–En yields the highest layer-wise average across all experimental settings, likely due to the shared Latin-based script, reinforcing prior findings on the role of script similarity in enhancing cross-lingual transfer~\citep{sindhujan-etal-2025-llms}.

The overall results from ALOPE indicate a decrease in correlation scores beyond the mid-Transformer layer (TL-16) for most models and language pairs. This aligns with the observation of~\citet{kargaran2024mexamultilingualevaluationenglishcentric}, that as embeddings for low-resource languages when progressed through the mid-Transformer layers, they became more aligned with the primary language of the LLMs. This improved alignment likely contributes to better cross-lingual representation at intermediate layers (TL-11, TL-7) after the mid-layers, and may also explain the decreased performance observed in the lower-level Transformer layers (TL-20, and TL-24), where such alignment has not yet stabilized.

% \vspace{-10pt}
\subsubsection{Model-wise analysis}\label{modelwise_analysis}
% \vspace{-4pt}
The best-performing models under the ALOPE framework consistently belonged to the LLaMA family. Notably, the LLaMA 3.2 model yielded the highest Spearman correlation scores for six out of eight language pairs: En to \{Gu, Hi, Mr, Ta\} and \{Et, Ne\} to En. LLaMA 3.1 demonstrated the best performance for En–Te, while LLaMA 2–7B outperformed other models on Si–En. These results highlight the effectiveness of LLaMA-based models for cross-lingual tasks over other open-source models.

Furthermore, statistical analysis using the Williams significance test revealed that although LLaMA 3.2 did not yield the absolute highest correlation scores for En–Te or Si–En, its performance at TL–7 and TL–11 was statistically indistinguishable from the best-performing models (as detailed in Table~\ref{tab:best-rh-layers_zero-shot}). Notably, LLaMA 3.2 attained this level of performance despite having the smallest parameter size (3 billion) among all the models evaluated in this study. This suggests that model size alone is not the sole determinant of the performance of a cross-lingual task such as QE and highlights the efficiency and adaptability of LLaMA 3.2 within resource-constrained environments. The Aya model also demonstrated correlation scores comparable to LLaMA models, with one notable exception, when the ALOPE experimented with the final Transformer layer (TL-1), its performance exhibited a significant decrease. The possible reason for this could be that the final layer of Aya is less aligned with quality estimation signals, whereas its intermediate representations still capture useful features. Overall, all the LLMs demonstrated improved performance when using ALOPE compared to SIFT, when the regression heads were attached to the optimal Transformer layers. 

\vspace{-10pt}
\subsection{Dynamic weighting \& Multi-head regression for QE}
\begin{figure*}[t]
\centering
\includegraphics[width=0.99\textwidth, keepaspectratio]{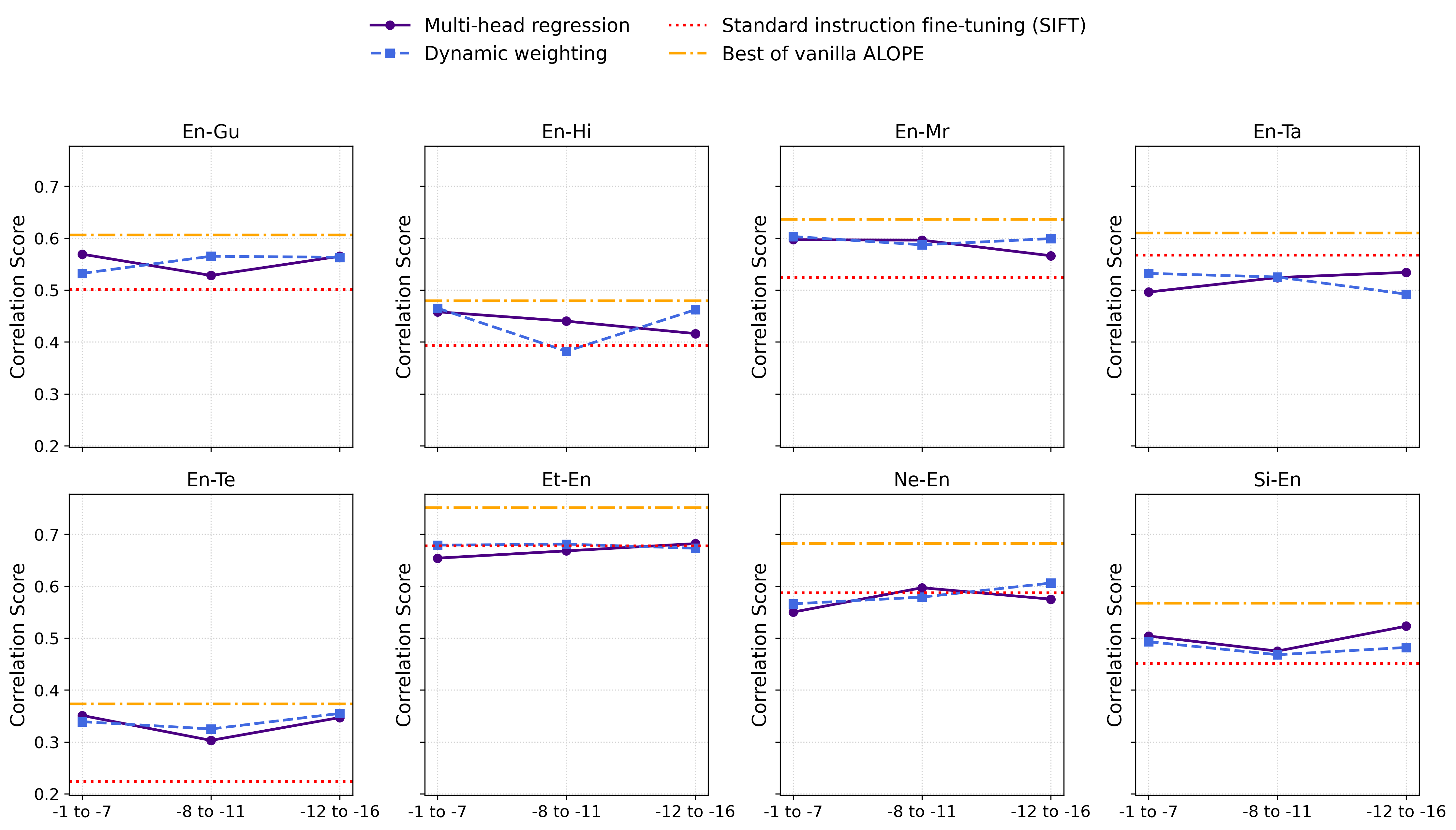}
\caption{\small{LLaMA3.2-3B results for dynamic-weighting and multi-head regression compared against standard instruction fine-tuning results and best of Vanilla ALOPE. Y-axis represents the correlation score and the X-axis represents the selected range of Transformer layers for dynamic weighting and multi-head regression experiments.} }
\label{fig:dynamic_combined_loss} % Give a unique label
\end{figure*}

% \vspace{-2pt}
To further evaluate the benefits of leveraging multiple Transformer layers, dynamic weighting and multi-head regression experiments were conducted using the LLaMA 3.2-3B model, which had previously demonstrated the most consistent performance across language pairs (Section~\ref{layerwise_analysis}). The experiments focused on three Transformer layer groupings: TL-1 to TL-7, TL-8 to TL-11, and TL-12 to TL-16. As illustrated in Figure~\ref{fig:dynamic_combined_loss}, both approaches yielded improvements over SIFT for En-\{Gu, Mr, Te\} and Si–En.  For the En-Hi language pair, the multi-head regression approach outperformed SIFT across all three layer groupings, whereas dynamic weighting exhibited a slight performance drop in the TL-8 to TL-11 configuration. In the case of Et-En and Ne-En, both multi-layer strategies delivered comparable performance to SIFT, with only marginal deviations observed. However, the En-Ta pair continued to yield suboptimal results under both approaches, resembling its weaker performance trends observed in previous experiments. This underscores the unique challenges associated with En-Ta in cross-lingual QE tasks.

In Figure~\ref{fig:dynamic_combined_loss}, Vanilla ALOPE represents the best correlation scores obtained through layer-specific adaptation with LLaMA3.2-3B model, as detailed in Section~\ref{layerwise_analysis}. When comparing the performance of multi-layer approaches with Vanilla ALOPE’s layer-specific adaptation, Vanilla ALOPE generally achieves higher correlation scores. Nevertheless, both multi-layer approaches demonstrate improvements over SIFT for the majority of the language pairs. These results highlight that multi-layer methods can also effectively leverage complementary information from multiple Transformer layers, making them valuable alternatives for efficient LLM-based translation quality estimation.

\vspace*{-5pt}

% \vspace{-10pt}
% \subsection{Multi-layer regression}

\subsection{Comparative analysis: ALOPE \textit{vs.} QE frameworks \& Monolingual setting}

\begin{table}[t]
\centering
\small
\begin{NiceTabular}{|l|ccccc:ccc|}
\toprule

\textbf{Method} & \textbf{En-Gu} & \textbf{En-Hi} & \textbf{En-Mr} & \textbf{En-Ta} & \textbf{En-Te} & \textbf{Et-En} & \textbf{Ne-En} & \textbf{Si-En} \\ 
\midrule
ALOPE\textsuperscript{*} & 0.606 & 0.479 & \textbf{0.636} & 0.610 & \textbf{0.388} & 0.751 & 0.682 & 0.573 \\ 
SIFT-LLMs\textsuperscript{*} & 0.555 & 0.393 & 0.530 & 0.586 & 0.290 & 0.728 & 0.618 & 0.558 \\ 
TransQuest\textsuperscript{*} & 0.630 & 0.478 & 0.606 & 0.603 & 0.358 & 0.760 & 0.718 & 0.579 \\ 
CometKiwi & \textbf{0.637} & \textbf{0.546} & 0.635 & \textbf{0.616} & 0.338 & \textbf{0.860} & \textbf{0.789} & \textbf{0.703} \\ 
\bottomrule
\end{NiceTabular}
\caption{\small{Comparison of highest Spearman correlation scores across ALOPE, standard instruction fine-tuning (SIFT) with LLMs, and encoder-based QE baselines—TransQuest and CometKiwi. Methods marked with (\textsuperscript{*}) are trained using the datasets described in Appendix~\ref{app:dataset}. The CometKiwi results are from the publicly available model pre-trained on over 940K QE data samples~\citep{rei-etal-2023-scaling}.
}}
\label{tab:best_comparison}
\end{table}
\vspace{-5pt}

The ALOPE approach demonstrated consistent improvements in correlation scores over SIFT across all language pairs and achieved results comparable to those of established pre-trained encoder-based QE frameworks, as shown in Table~\ref{tab:best_comparison}. Notably, for the En-Mr and En-Te language pairs, ALOPE even surpassed the SOTA performance of these frameworks. While pre-trained encoder-based models such as TransQuest (InfoXLM-Large) and CometKiwi (XLM-R-XL) continue to show strong performance, they require dedicated training pipelines and a significant amount of data for fine-tuning. In contrast, ALOPE offers a flexible and modular solution that can be applied to pre-deployed LLMs without extensive reconfiguration, making it a practical solution for performing QE. 

Despite being based on LLMs, ALOPE demonstrates favourable memory efficiency when compared to encoder-based QE frameworks. Although LLMs are often considered more resource-intensive, our implementation of ALOPE—augmented with LoRA-adapted lightweight regression heads exhibits competitive GPU memory usage, as detailed in Appendix~\ref{app:gpu_memory}. This indicates that ALOPE can be deployed with overheads comparable to, or even lower than, those of established encoder-based models, highlighting its practicality in resource-constrained environments. Finally, it is also scalable to a unified evaluation and correction framework where an additional causal head can provide translation error reasoning-based corrections or perform automatic post-editing.

To evaluate the generalizability of ALOPE for regression tasks beyond cross-lingual quality estimation, we conducted an ablation study, which is detailed in the Appendix~\ref{app:mono_vs_multi}, observing its performance on a monolingual regression task. Results demonstrated that ALOPE consistently outperforms standard instruction fine-tuning and baseline scores in the monolingual setting, achieving the highest scores across evaluated languages- English, Spanish and Arabic. These findings highlight ALOPE's ability to effectively identify optimal transformer layers for enhanced performance in both monolingual and cross-lingual settings.
% \vspace{-15pt}

\section{Conclusion}\label{sec:conclusion}
% \vspace{-7pt}
This paper introduced ALOPE, a novel framework for improving segment-level translation quality estimation using LLMs. By integrating regression heads with LoRA and strategically leveraging intermediate Transformer layers, ALOPE outperforms standard instruction fine-tuning results of LLMs and achieves results on par with, or better than, encoder-based QE frameworks. Our experiments demonstrate that intermediate Transformer layers, particularly TL-7, yield more effective cross-lingual representations for QE, while performance tends to decrease in layers beyond the mid-range. Notably, LLaMA 3.2 emerged as the most effective model across language pairs, despite having the smallest parameter size, underscoring that parameter count is not the sole indicator of QE effectiveness. Furthermore, early stabilization of the performance was observed when English served as the target language in translation, especially around mid-layer depths. Additionally, the framework introduces dynamic weighting and multi-head regression approaches, which leverage a selective set of multiple Transformer layers to perform QE. ALOPE demonstrates competitive GPU efficiency, reinforcing its practicality as a flexible and scalable solution for deploying LLMs in resource-constrained settings to perform QE. We also envision this framework to be extended for reasoning over MT errors informed by a regression-head predicted DA, providing more reliability to error reasoning and MT assessment.

\bibliography{colm2025_conference}
\bibliographystyle{colm2025_conference}
\clearpage
\appendix

\section{Dataset} \label{app:dataset}

\begin{table}[ht!]
\centering
\small % Reduces the font size
\setlength{\tabcolsep}{12pt} % Adjust column spacing
\begin{tabular}{@{}lcccccccc@{}}
\toprule
& \textbf{En-Gu} & \textbf{En-Hi} & \textbf{En-Mr} & \textbf{En-Ta} & \textbf{En-Te} & \textbf{Et-En} & \textbf{Ne-En} & \textbf{Si-En} \\ \midrule
\textbf{Train} & 7000 & 7000 & 26000 & 7000 & 7000 & 7000 & 7000 & 7000 \\
\textbf{Test}  & 1000 & 1000 & 699   & 1000 & 1000 & 1000 & 1000 & 1000 \\ \bottomrule
\end{tabular}
\caption{Dataset splits of QE dataset utilized in our study. Experiments were conducted with 8 low-resource language pairs to evaluate model performance.}
\label{tab:datasets_crosslingual}
\end{table}
% \clearpage

\section{Results of standard instruction fine-tuned LLMs (SIFT)} \label{app:finetune}

\begin{table*}[htbp]
% \scriptsize
\setlength{\tabcolsep}{4pt} % Adjust column spacing
\centering
\resizebox{\textwidth}{!}{%
\begin{tabular}{l|ccccccccc}
\toprule
\textbf{Models} & \textbf{En-Gu} & \textbf{En-Hi} & \textbf{En-Mr} & \textbf{En-Ta} & \textbf{En-Te} & \textbf{Et-En} & \textbf{Ne-En} & \textbf{Si-En} & \textbf{\textit{Avg}} \\
\midrule

LLaMA2-7B & 0.454 & 0.319 & 0.466 & 0.449 & 0.266 & 0.591 & 0.478 & 0.374 & 0.425 \\

LLaMA3.1-8B & 0.555 & 0.368 & \textbf{0.530} & 0.586 & 0.233 & \textbf{0.728} & \textbf{0.618} & \textbf{0.558} & 0.522 \\
LLaMA3.2-3B & 0.501 & \textbf{0.393} & 0.524 & \textbf{0.567} & 0.224 & 0.678 & 0.587 & 0.451 & 0.491 \\
Aya-expanse-8B & \textbf{0.528} & 0.388 & 0.515 & 0.496 & \textbf{0.290} & 0.605 & 0.568 & 0.384 & 0.472 \\

\textbf{\textit{Avg}} & 0.510 & 0.367 & 0.509 & 0.525 & 0.253 & 0.650 & 0.563 & 0.442 &  \\

\bottomrule
\end{tabular}}
\caption{Spearman correlation score obtained with standard instruction fine-tuning. Bolded values denote the highest Spearman score obtained for each language pair}
\label{tab:multi-language}
\end{table*}

\section{GPU memory utilization} \label{app:gpu_memory}

 \begin{figure*}[ht]
 \centering
 \includegraphics[width=0.98\textwidth, keepaspectratio]{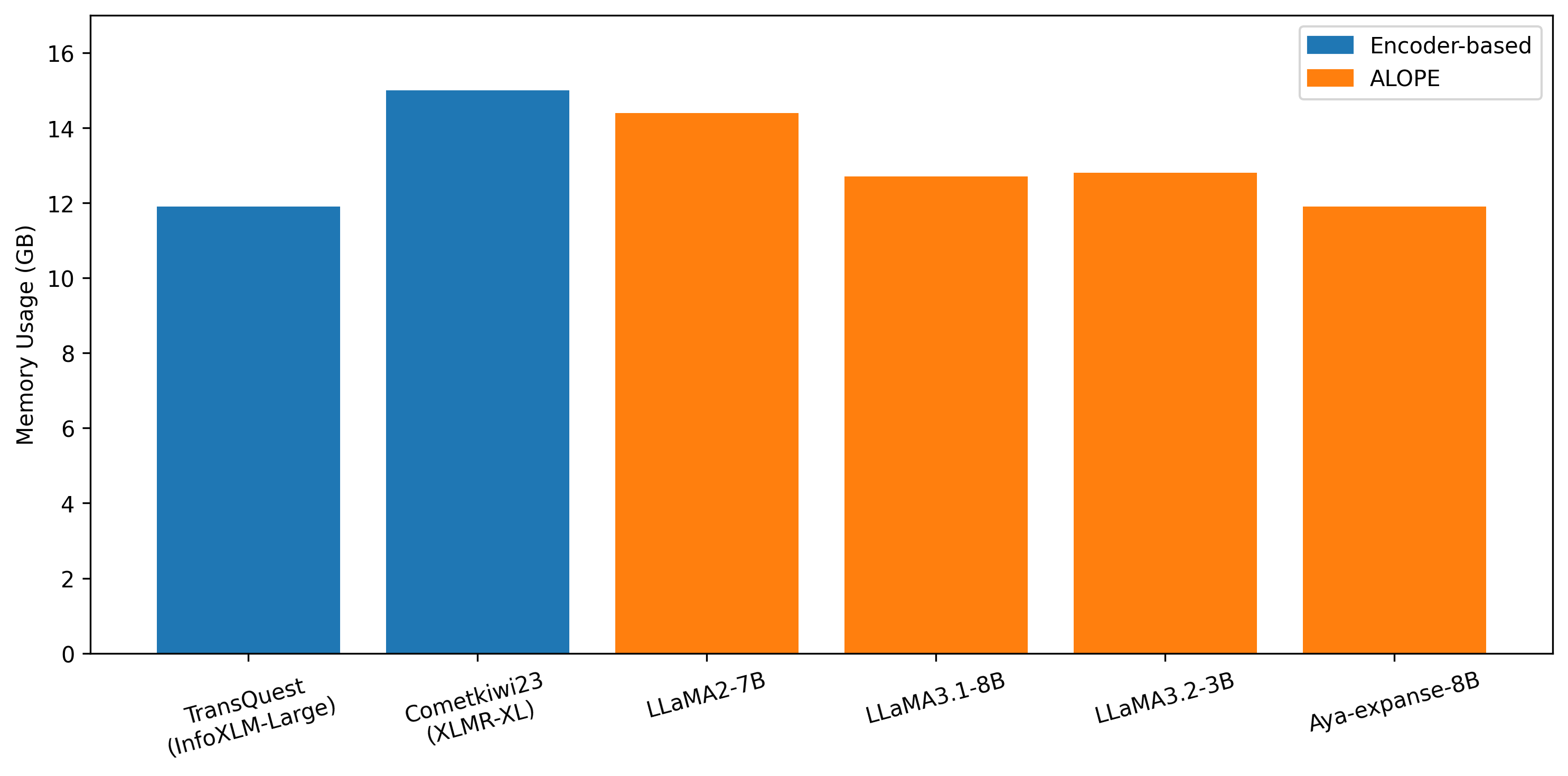}
\caption{ GPU memory utilization comparison between encoder-based QE models and various LLM-based ALOPE configurations.}
\label{fig:memory} % Give a unique label
\end{figure*}

\clearpage

\section{Language pair-wise performance across different Transformer layers} \label{app:all-lang-layerwise-performance}

\begin{figure*}[htbp]
\centering
\includegraphics[width=0.88\textwidth, keepaspectratio]{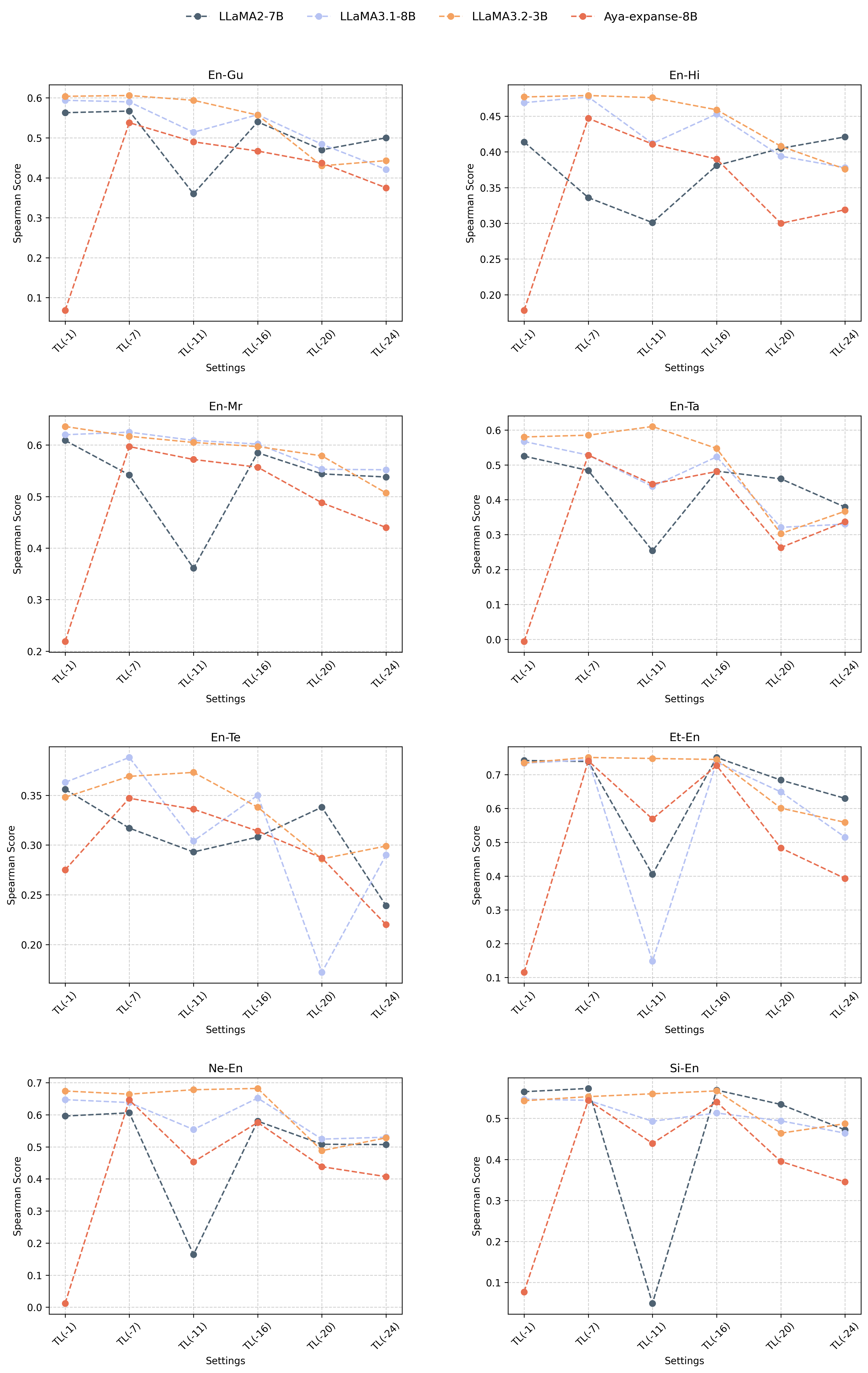}
\caption{\small{ Spearman correlation scores obtained across different Transformer layers with ALOPE layer-specific regression experiments. } }
\label{fig:LP-wise-performance} % Give a unique label
\end{figure*}

\clearpage

\section{Assessing the generalization of ALOPE across regression tasks: An ablation study}\label{app:mono_vs_multi}

 \begin{figure*}[ht]
 \centering
 \includegraphics[width=0.98\textwidth, keepaspectratio]{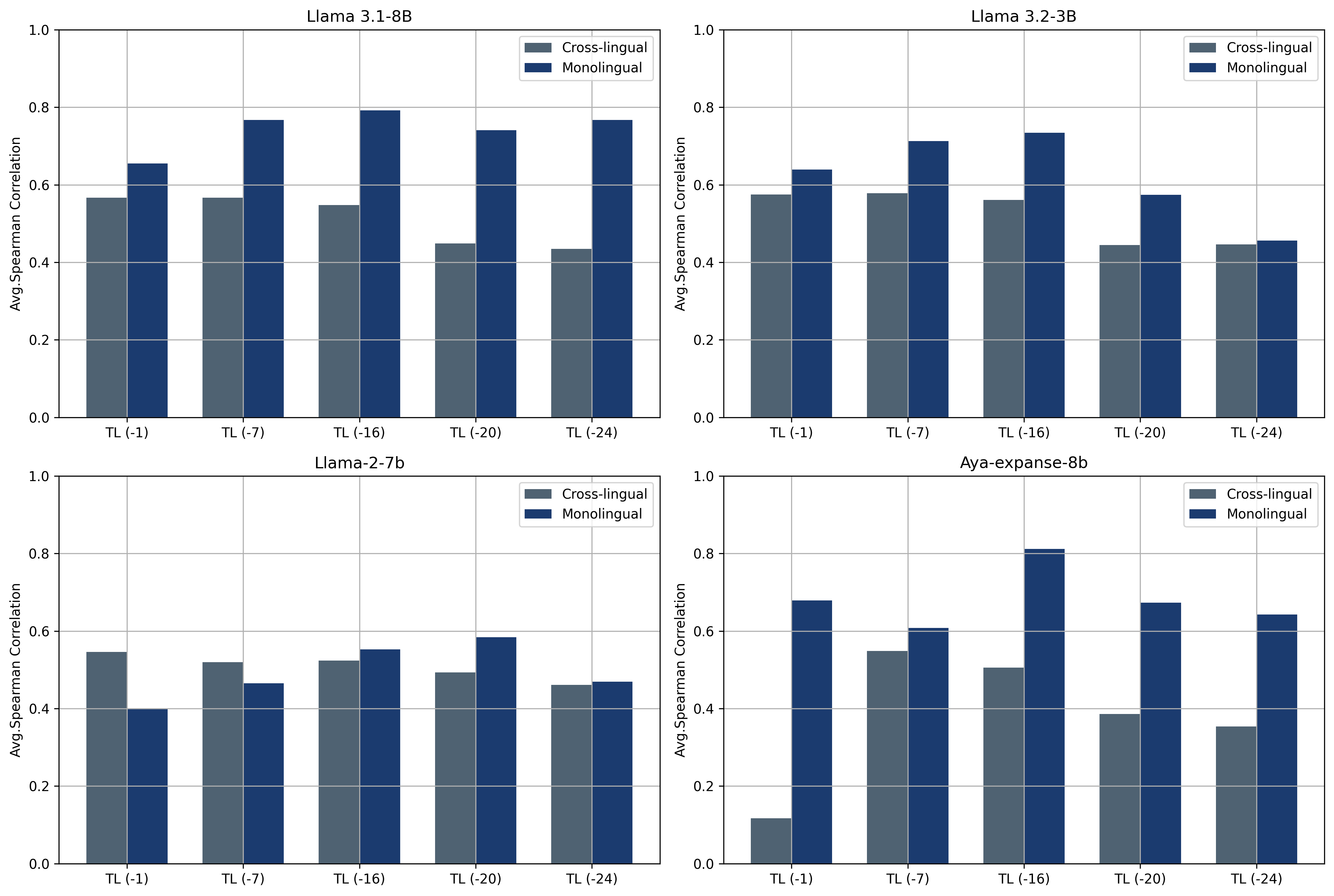}
\caption{ Model-wise average Spearman correlation scores obtained with layer-specific ALOPE approach, comparing performance on monolingual versus cross-lingual regression tasks. }
\label{fig:mono_vs_cross} % Give a unique label
\end{figure*}

The ALOPE framework consistently outperformed standard instruction fine-tuned LLMs in the cross-lingual regression task of quality estimation across all language pairs and delivered results comparable to leading encoder-based QE models (Table~\ref{tab:best_comparison}). For En-Mr and En-Te, it even exceeded state-of-the-art performance. To better understand whether the performance gains observed with ALOPE are specific to cross-lingual QE or indicative of broader regression capabilities, we conducted an ablation study using a monolingual regression task.

For the comparative analysis, we utilized the SemEval-2018 shared task dataset~\citep{mohammad-etal-2018-semeval}, which focuses on predicting emotion intensity scores, real-valued outputs ranging from 0 to 1 based on tweet content. The dataset spans four emotion categories and comprises monolingual data in English, Spanish, and Arabic. SemEval was selected for its structural similarity to the QE dataset, as both tasks involve predicting a continuous score based on the text input. However, the key distinction lies in language composition: SemEval is monolingual, whereas QE contains source-target translation data, which is cross-lingual. This contrast makes SemEval a suitable benchmark for evaluating ALOPE's performance in monolingual versus cross-lingual regression settings.

Experiments were conducted using the same set of LLMs within the ALOPE framework, demonstrating its flexibility in adapting to monolingual regression tasks.  As shown in the Table ~\ref{tab:emotion}, all three languages obtained the best score with ALOPE for emotion intensity score prediction showing improved correlation score over the baseline, \textit{i.e.}, best score obtained for the SemEval-2018 shared task~\citep{mohammad-etal-2018-semeval} and standard instruction fine-tuned results of LLMs. This shows the effectiveness of ALOPE for LLM-based regression tasks.

Although direct comparison between monolingual and cross-lingual tasks is not feasible due to their differing objectives, we analyzed model-wise average Spearman correlation scores across selected Transformer layers (see Figure~\ref{fig:mono_vs_cross}). This graph shows that the monolingual regression task consistently achieves higher average correlation scores across most of the Transformer layers, including the lower-level layers (TL:-20 and -24). In contrast, cross-lingual QE tasks show improved performance predominantly in the intermediate layers beyond the mid-level. These findings reinforce ALOPE's effectiveness in identifying and leveraging the most suitable Transformer layers for cross-lingual quality estimation, thereby enhancing performance in low-resource settings.  

Building on this analysis, the ALOPE framework demonstrates strong adaptability and consistently outperforms standard LLM baselines across both monolingual and cross-lingual regression tasks, highlighting its generalizability and effectiveness in identifying optimal transformer layers for varied language settings.

 \begin{figure*}[t]
 \centering
 \includegraphics[width=0.9\textwidth, keepaspectratio]{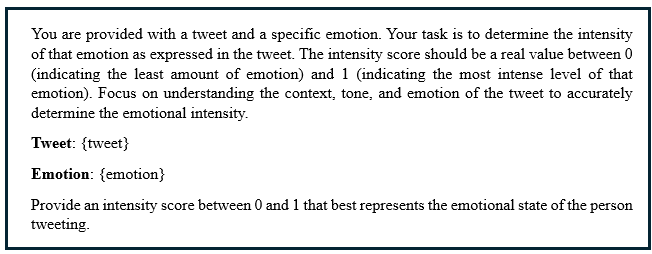}
\caption{ Prompt for emotion intensity prediction}
\label{fig:emotion-intensity} % Give a unique label
\end{figure*}

\begin{table}[t]
\centering
\small % Reduces the font size
\setlength{\tabcolsep}{12pt} % Adjust column spacing
\begin{tabular}{@{}lccc@{}}
\toprule
& \textbf{English} & \textbf{Spanish} & \textbf{Arabic} \\ \midrule
\textbf{Train } & 7102 & 4544 & 3376 \\
\textbf{Test }  & 4068 & 2616 & 1563 \\ \bottomrule
\end{tabular}
\caption{Data count of each split utilized for emotion intensity score prediction.}
\label{tab:data_emotion}
\end{table}

\begin{table*}[t]
\small
\setlength{\tabcolsep}{4pt} % Adjust column spacing
\centering
\begin{adjustbox}{max width=\textwidth}
% \begin{NiceTabular}{l|lccccccccc}
\begin{NiceTabular}{|l|p{2.5cm}|cc:cc:cc|}

\toprule
 & \textbf{Model} & \multicolumn{2}{c}{\textbf{English}} & \multicolumn{2}{c}{\textbf{Spanish}} & \multicolumn{2}{c}{\textbf{Arabic}}  \\
 & & \textbf{$\rho$} & \textbf{$r$} & \textbf{$\rho$} & \textbf{$r$} & \textbf{$\rho$} & \textbf{$r$}  \\
\midrule

\multirow{5}{*}{\rotatebox[origin=c]{90}{ \textbf{SIFT} }}
& llama-2-7b & 0.766 & 0.767 & 0.775 & 0.778 & 0.481 & 0.486  \\

& llama 3.1-8B & 0.794 & 0.792 & 0.797 & 0.798 & 0.662 & 0.670  \\
& llama 3.2-3B & 0.777 & 0.775 & 0.755 & 0.752 & 0.546 & 0.558  \\
& aya-expanse-8b & 0.779 & 0.782 & 0.807 & 0.807 & 0.721 & 0.727 \\
\\
& \textit{\textbf{Avg}} & \textit{0.779} & - & \textit{0.783} & - & \textit{0.602} & -   \\
\midrule

\multirow{5}{*}{\rotatebox[origin=c]{90}{ \textbf{TL (-1)} }}
& llama-2-7b & 0.755 & 0.759 & 0.321 & 0.320 & 0.123 & 0.116  \\

& llama 3.1-8B & 0.786 & 0.788 & 0.709 & 0.709 & 0.471 & 0.480  \\
& llama 3.2-3B & 0.773 & 0.771 & 0.706 & 0.705 & 0.438 & 0.440  \\
& aya-expanse-8b & 0.767 & 0.771 & 0.760 & 0.760 & 0.510 & 0.516  \\
\\
& \textit{\textbf{Avg}} & \textit{0.770} & - & \textit{0.624} & - & \textit{0.385} & -   \\
\midrule

\multirow{5}{*}{\rotatebox[origin=c]{90}{ \textbf{TL (-7)} }}
& llama-2-7b & 0.747 & 0.748 & 0.410 & 0.407 & 0.239 & 0.238  \\

& llama 3.1-8B & 0.823 & 0.825 & 0.812 & 0.815 & 0.665 & 0.670  \\
& llama 3.2-3B & 0.801 & 0.803 & 0.787 & 0.791 & 0.553 & 0.556  \\
& aya-expanse-8b & 0.686 & 0.683 & 0.657 & 0.662 & 0.482 & 0.490  \\ \\
& \textit{\textbf{Avg}} & \textit{0.764} & - & \textit{0.666} & - & \textit{0.485} & -  \\
\midrule

\multirow{5}{*}{\rotatebox[origin=c]{90}{ \textbf{TL (-16)} }}
& llama-2-7b & 0.801 & 0.804 & 0.748 & 0.750 & 0.111 & 0.104  \\

& llama 3.1-8B & 0.830 & 0.829 & \textbf{0.840} & 0.825 & 0.707 & 0.712  \\
& llama 3.2-3B & 0.802 & 0.804 & 0.791 & 0.794 & 0.611 & 0.611  \\
& aya-expanse-8b & 0.826 & 0.830 & 0.839 & 0.841\textsuperscript{*} & \textbf{0.769} & 0.779\textsuperscript{*}  \\
\\
& \textit{\textbf{Avg}} & \textit{0.814} & - & \textit{0.804 }& - & \textit{0.549} & -   \\
\midrule
\multirow{5}{*}{\rotatebox[origin=c]{90}{ \textbf{TL (-20)} }}
& llama-2-7b & 0.821 & 0.823 & 0.753 & 0.756 & 0.177 & 0.186  \\
& llama 3.1-8B & \textbf{0.831} & 0.831\textsuperscript{*} & 0.833 & 0.835 & 0.559 & 0.567  \\
& llama 3.2-3B & 0.801 & 0.802 & 0.741 & 0.744 & 0.178 & 0.166  \\
& aya-expanse-8b & 0.784 & 0.788 & 0.757 & 0.758 & 0.480 & 0.491  \\ \\
& \textit{\textbf{Avg}} & \textit{0.809} & - & \textit{0.771} & - & \textit{0.348} & -  \\
\midrule
\multirow{5}{*}{\rotatebox[origin=c]{90}{ \textbf{TL (-24)} }}
& llama-2-7b & 0.745 & 0.742 & 0.381 & 0.375 & 0.281 & 0.288  \\
& llama 3.1-8B & 0.806 & 0.808 & 0.787 & 0.790 & 0.708 & 0.717  \\
& llama 3.2-3B & 0.635 & 0.632 & 0.311 & 0.307 & 0.420 & 0.420  \\
& aya-expanse-8b & 0.733 & 0.731 & 0.619 & 0.615 & 0.576 & 0.589  \\ \\
& \textit{\textbf{Avg}} & \textit{0.730} & - & \textit{0.525} & - & \textit{0.496} & -  \\
\midrule

& SemEval Task Best & - & 0.799 & - & 0.738 & - & 0.685   \\

\bottomrule
\end{NiceTabular}
\end{adjustbox}
\caption{Spearman's ($\rho$) and Pearson's ($r$) correlation scores obtained with ALOPE framework for the monolingual regression task of emotion intensity prediction across multiple language pairs. The last row in the table shows the \textit{best} Pearson correlation score obtained in SemEval-2018 for the emotion-intensity score prediction task \citep{mohammad-etal-2018-semeval}. The bolded values show the highest Spearman score obtained and asterisks (*) indicate the highest Pearson score obtained.}
\label{tab:emotion}
\end{table*}

\clearpage

\end{document}